\documentclass{article}

\usepackage[preprint]{neurips_2026}

\usepackage[utf8]{inputenc} 
\usepackage[T1]{fontenc}    
\usepackage{hyperref}       
\hypersetup{
    colorlinks,
    allcolors={blue!80!},
}
\usepackage{url}            
\usepackage{booktabs}       
\usepackage{amsfonts}       
\usepackage{nicefrac}       
\usepackage{microtype}      
\usepackage{xcolor}         

\usepackage{csquotes}
\usepackage{xspace}
\usepackage{enumitem}
\usepackage{amsmath, bm, bbm}
\usepackage{mathtools}
\usepackage{wrapfig}
\usepackage{caption}
\usepackage{multirow}
\usepackage{booktabs}
\usepackage[table]{xcolor}
\usepackage[noend]{algpseudocode}
\usepackage{algorithm}
\usepackage{eqparbox}
\usepackage{amssymb}
\usepackage[most]{tcolorbox}
\usepackage{rotating}
\usepackage{soul}
\usepackage[dvipsnames]{xcolor}
\setul{0.5ex}{}
\usepackage{graphicx}
\usepackage{lipsum}

\title{\smash{\raisebox{-0.27em}{\includegraphics[height=1.2em]{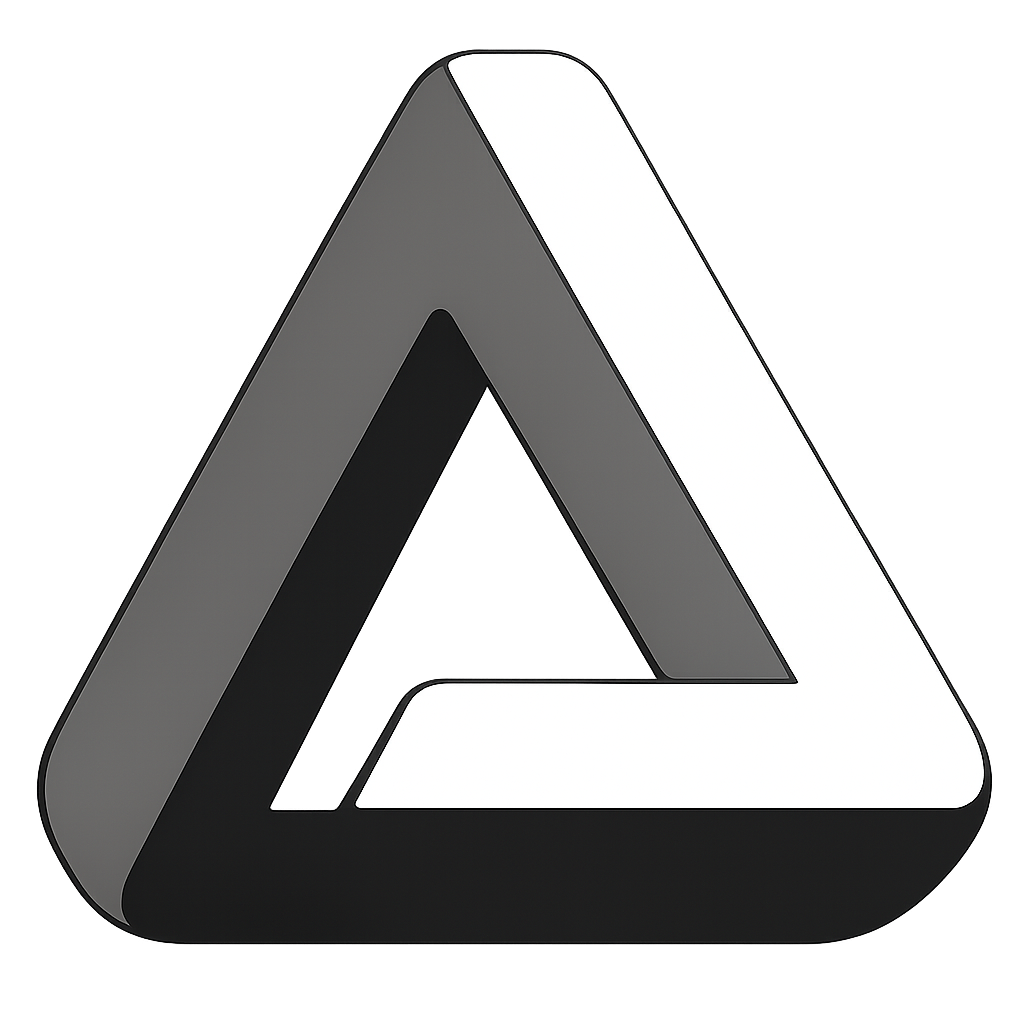}}} DeltaPrompts: Escaping the Zero-Delta Trap in Multimodal Distillation}

\author{
Jaehun Jung \hspace{.3cm}
Hyunwoo Kim \hspace{.3cm}
Brandon Cui \hspace{.3cm}
Ximing Lu \hspace{.3cm} \\
\textbf{David Acuna} \hspace{.3cm}
\textbf{Prithviraj Ammanabrolu} \hspace{.3cm}
\textbf{Yejin Choi} \\
\small{\quad}\\
NVIDIA Research
}

\newcommand{\eg}{\textit{e.g.,} }
\newcommand{\ie}{\textit{i.e.,} }

\newcommand{\dataset}{\textsc{DeltaPrompts}\xspace}

\algnewcommand\Break{\textbf{break}}
\DeclareMathOperator*{\randomSample}{random-sample}
\DeclareMathOperator*{\kMeans}{K-means}
\begin{document}

\maketitle

\begin{abstract}
Distillation enables compact Vision-Language Models (VLMs) to obtain strong reasoning capabilities, yet the prompts driving this process are typically chosen via simple heuristics or aggregated from off-the-shelf datasets. We reveal a critical inefficiency in this approach: up to 69\% of the prompts in standard chart / document reasoning datasets are effectively \textit{zero-delta}, meaning the teacher and student already induce the exact same answer distribution. Training on these prompts provides minimal learning signal, causing student improvement to rapidly saturate regardless of data scale. To escape the zero-delta trap, we return to first principles: distillation fundamentally minimizes distributional divergence, and thus a prompt is valuable only if it exposes a functional capability gap between the teacher and student. We quantify this gap through \textit{answer divergence} ($\Delta$), demonstrating that non-zero divergence is critical for effective scaling. Building on this insight, we propose a staged synthesis pipeline that repurposes existing datasets as seeds, actively targeting student failure modes to produce better prompts. The result is \textbf{\dataset}, a diverse dataset of 200k synthetic, high-divergence reasoning problems. We evaluate \dataset across three distinct settings: on-policy distillation with the target teacher-student pair, transfer to a novel model family without regenerating the data, and off-policy fine-tuning of a non-reasoning model. Across all scenarios, \dataset drives substantial gains, yielding up to 15\% relative improvement even on top of a highly-optimized reasoning model (\eg Qwen3-VL-8B-Thinking)---averaged over 10 benchmarks spanning chart, document and perception-centric reasoning.
\end{abstract}
\section{Introduction}
Distillation plays a pivotal role in modern Vision-Language Models, offering compact models an efficient path to learn from a stronger teacher \cite{qwen3-vl, glm}. A growing body of work has advanced the algorithmic foundations of this process---from stabilizing divergence objectives \cite{minillm, distillm, rethinking-kl} to on-policy variants that mitigate distribution shift \cite{gkd, opsd, opd-thinking-machines}---and has seen particular success on reasoning-heavy tasks. But across these advancements, comparatively little attention has been paid to the \textit{prompts} themselves: in practice, reasoning problems are often collected from a standard array of datasets \cite{finevision, bee, mmfinereason, tinyeye}, either selected by researchers or via heuristics such as difficulty. It remains an open question whether the off-the-shelf prompts are indeed suitable for closing the capability gap, and if not, how to systematically improve them.

In this work, we study what makes a prompt effective for multimodal distillation—and how to construct better ones at scale. We argue that a prompt is valuable only insofar as it exposes a \textit{capability gap} between teacher and student---when the two models induce effectively the same answer distribution on the input, the distillation signal reveals little about what the student lacks. From this perspective, the quality of a distillation corpus depends not on how many raw samples it contains, but on \textit{how often it surfaces the prompts where the teacher and student actually disagree.}

\begin{figure}[t]
  \centering
  \vspace{-5pt}
  \includegraphics[width=.96\textwidth]{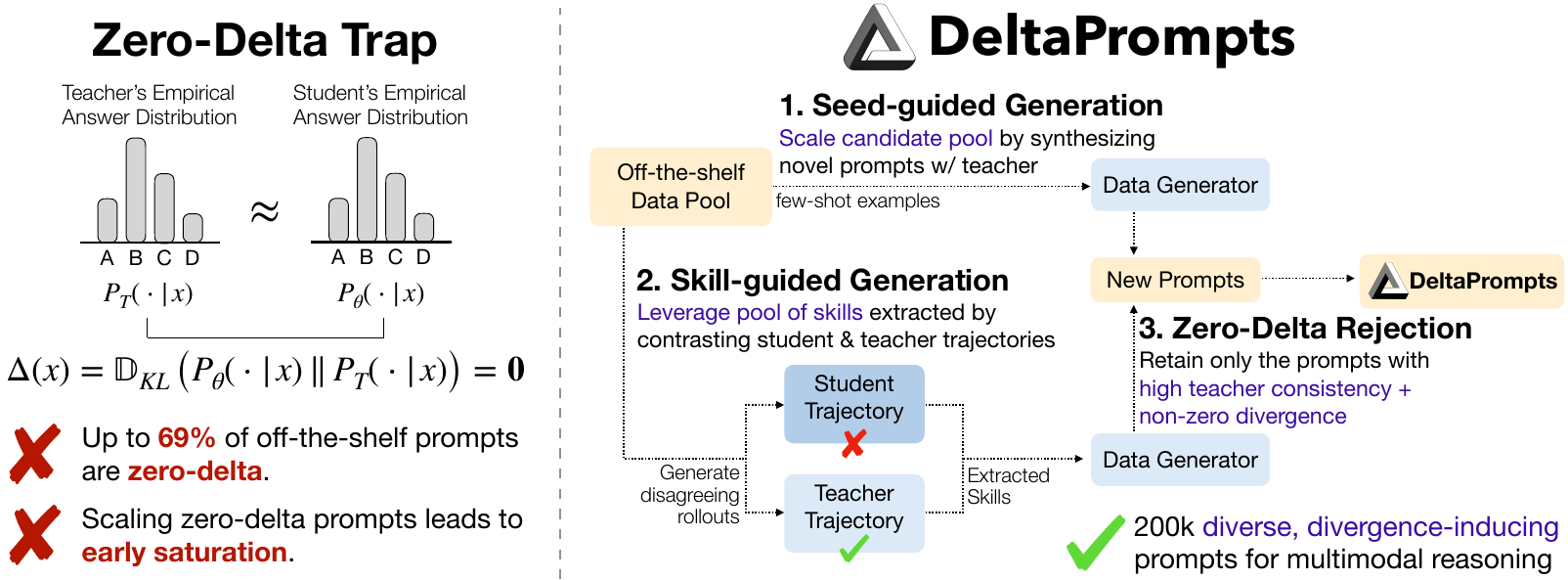} vb 
  \caption{\textbf{Overview of \dataset}. We first identify the \textit{zero-delta trap}: for majority of the prompts in chart \& document reasoning datasets, the teacher and student already produce an identical answer distribution. We show that distillation on \textbf{these zero-delta prompts reaches early saturation in student performance} (\S \ref{sec:impact_of_answer_divergence}). We then synthesize \textbf{\dataset, leveraging teacher model itself to find the divergence-inducing prompts, and using off-the-shelf data as seeds rather than a direct training resource} (\S \ref{sec:synthesizing_deltaprompts}).}
  \vspace{-10pt} 
  \label{fig:placeholder}
\end{figure}

To make this idea practical, we measure the divergence between the discrete answer distribution induced by the two models given a prompt, which we call \textbf{answer divergence} ($\Delta$). Despite its simplicity---sidestepping the noisiness and cost of sequence-level divergence---answer divergence proves critical for effective distillation: training on \textit{delta prompts} with non-zero $\Delta$ continues to improve the student, while \textit{zero-delta prompts} saturate rapidly regardless of data scale. Importantly, our analysis reveals that existing datasets fall short precisely on answer divergence---\textbf{up to 69\% of the prompts in standard chart / document reasoning datasets are effectively zero-delta}, \ie the student and teacher produce the exact same answer distribution. The remaining 31\% exhibit significantly lower diversity than the original pool, making naive filtering an unreliable solution.

Motivated by these observations, we repurpose existing datasets not as a direct training source, but as seeds for synthesizing better prompts. Our key insight is that unlike fixed data curation, synthesis allows us to actively search for the properties that matter in distillation---\textbf{non-zero divergence and broad skill coverage}. We materialize this idea into a staged pipeline that (1) expands the prompt pool via seed-guided synthesis, and (2) augments specific reasoning skills that student fails to exhibit, all with the teacher model itself as the data generator. The resulting prompts exhibit both higher answer divergence and greater diversity than the existing datasets they originated from, even after aggressive filtering for quality.

We apply our pipeline to chart \& document reasoning and high-resolution, perception-centric reasoning---domains where high-quality reasoning problems are particularly sparse---and produce \textbf{\dataset}, a dataset of 200k synthetic, divergence-inducing prompts. With on-policy distillation, \dataset yields up to 15\% relative improvement across benchmarks, even on top of an already highly-optimized reasoning model (\eg Qwen3-VL-8B-Thinking \cite{qwen3-vl}). These gains transfer to an entirely different model family (\eg GLM-4 \cite{glm}) without regenerating the data, and extend to out-of-domain tasks such as STEM reasoning. \dataset also serves as an effective off-policy SFT resource, substantially outperforming off-the-shelf data mixtures. Taken together, our results suggest that the central bottleneck in multimodal distillation is not in aggregating more prompts or trajectories, but in constructing prompts where the teacher and student meaningfully disagree.
\section{What prompts are fit for distillation?} \label{sec:which_prompts_are_fit}
\subsection{Formulating Distillation Objective}
Given a teacher model $\pi_T$ and a student model $\pi_\theta$, knowledge distillation optimizes the student to minimize the divergence between the two models' output distributions conditioned on a prompt $x$:
\begin{equation}
    \mathcal{L}_{\text{distill}}(\theta) = \mathbb{E}_{x \sim \mathcal{D}} \left[ \mathbb{D} \left( \pi_T(\cdot | x) \parallel \pi_\theta(\cdot | x) \right) \right]
\end{equation}
where $\mathcal{D}$ is the prompt distribution and $\mathbb{D}$ is a divergence measure. The choice of divergence $\mathbb{D}$ and the sampling strategy for outputs give rise to different instantiations of this objective---for example, in supervised knowledge distillation \cite{distillbert, hinton-kd}, outputs are sampled from the teacher $\pi_T$ and the student is trained to minimize the forward KL, either by matching the token-level probabilities or by maximizing likelihood of teacher outputs \cite{sequence-kd, impossible-distillation}. Conversely, on-policy distillation (OPD) \cite{gkd} samples from the student's own distribution and minimizes the reverse KL:
\begin{equation}
    \mathcal{L}_{\text{OPD}}(\theta) = \mathbb{E}_{x \sim \mathcal{D}} \left[ \mathbb{E}_{\widehat{y} \sim \pi_\theta(\cdot | x)} \left[ \log \frac{\pi_\theta(\widehat{y}|x)}{\pi_T (\widehat{y} |x)}\right]\right]
\end{equation}
Beyond its original formulation, several additional improvements have been proposed to augment this objective, such as harnessing the student model itself as a teacher with privileged information \cite{opsd, opcd}, better decoding mechanism \cite{kd_with_mbr}, and auxiliary training signals \cite{entropy-aware-opd, self-distilled-rlvr}. We discuss further related works in \S \ref{app:related_works}.

Setting aside the algorithmic specifics, we observe that the learning signal in distillation is fundamentally driven by the distributional gap between teacher and student---\ie how the two models diverge in their outputs given an input $x$. From the perspective of dataset $\mathcal{D}$, the value of each prompt $x \in \mathcal{D}$ ultimately reduces to how much capability gap it induces between the two models. In principle, one could quantify this per-prompt gap by directly measuring the sequence-level divergence $\mathbb{D}(\pi_T(\cdot|x) ||\pi_\theta(\cdot|x))$. But in practice, this is both computationally intractable and highly noisy: the distributions are defined over variable-length sequences $\hat{y}$, where a trivial phrasing difference or an early token deviation can dominate the estimate of divergence.

\subsection{Answer Divergence $\Delta$} \label{sec:answer_divergence_delta}
Instead of comparing full output sequences, we propose answer divergence $\Delta$ over the discrete space of final answers. This reduces the support from variable-length token sequences to a much smaller set of equivalence classes over final answers, making the computation tractable while focusing on the functional disagreement between models—whether they arrive at different conclusions, not whether they phrase the same conclusion differently.

\paragraph{Defining answer divergence} Let $\mathcal{Y}$ denote the space of all possible variable-length sequences, and $\mathcal{A}$ denote the discrete space of possible final answers (\eg multiple-choice options, extracted scalar values, or open-ended answers). We define an extraction function $\mathcal{E}: \mathcal{Y} \rightarrow \mathcal{A}$ that maps a full sequence $y$ to its final answer $a$. We then naturally define the induced answer distribution as:
\begin{equation}\label{eq:induced_answer_distribution}
    P_\pi(a|x) = \sum_{y \in \mathcal{Y}: \,\mathcal{E}(y) = a} \pi(y|x) = \mathbb{E}_{y \sim \pi(\cdot|x)} \left[ \mathbb{I}(\mathcal{E}(y) = a)\right]
\end{equation}
The answer divergence $\Delta(x)$ can now be formulated by applying our chosen divergence measure $\mathbb{D}$ (\eg reverse KL) over the induced answer distributions of the two models:
\begin{equation}
    \Delta(x) = \mathbb{D}_{\text{KL}} \left( P_\theta(\cdot|x)||P_T(\cdot|x)\right)
\end{equation}
In practice, the exact marginalization over the sequence space in Eq. \ref{eq:induced_answer_distribution} is computationally intractable. We therefore approximate the true answer distribution by sampling $K=16$ trajectories from both the teacher and student for a given prompt $x$. To robustly handle surface-level variations in the answer space, we employ an LLM judge (Qwen3-32B) to evaluate the semantic equivalence of final answers across the $K$ samples. Grouping equivalent answers, we construct the empirical answer distributions, $\widehat{P}_T(a|x)$ and $\widehat{P}_\theta(a|x)$, from which we compute the empirical answer divergence $\widehat{\Delta}(x)$. In addition to handling surface-level variations, our LLM-judge based approach generalizes to open-ended or long-form answers, where exact string matching fails to capture semantic equivalence.

\begin{figure*}[t]
    \centering
    \includegraphics[width=.96\textwidth]{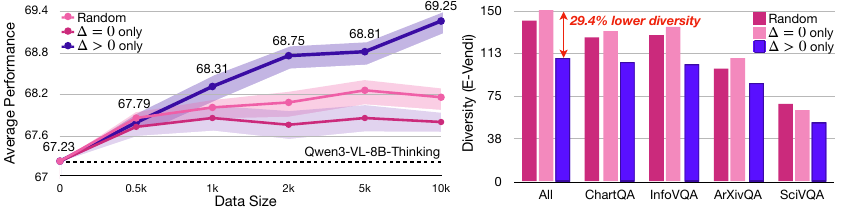}
    \caption{(Left) \textbf{Divergence is critical for effective OPD}. When controlling for diversity, delta prompts ($\Delta > 0$) improves with data size while random or zero-delta subsets meet early saturation. (Right) \textbf{Existing datasets trade off divergence against diversity}---filtering existing mixtures for delta prompts ($\Delta > 0$) reduces diversity by up to 29.4\% as measured by E-Vendi.}
    \vspace{-10pt}
    \label{fig:scaling_and_diversity}
\end{figure*}

\subsection{Impact of Answer Divergence on Distillation} \label{sec:impact_of_answer_divergence}
Our next step is to validate whether the answer divergence of the prompts actually leads to systematic differences in model performance. To this end, we run a series of controlled training runs using standard multimodal datasets for chart \& document reasoning. 

We first aggregate the train splits of off-the-shelf datasets---ChartQA, SciVQA, InfoVQA, arXivQA \cite{chartqa, scivqa, infovqa, arxivqa}---to create a mixture of 139k unique reasoning problems and measure their answer divergence. While controlling for the data scale ($N=0.5k, \cdots, 10k$), we sample 3 different types of subsets with unique divergence characteristics: (1) \textit{delta prompts} ($\Delta$ > 0 only), (2) \textit{zero-delta prompts}: ($\Delta = 0$ only), and (3) a random subset. During this process, we find that data diversity confounds the effect of answer divergence---\eg naive sampling of prompts with non-zero $\Delta$ significantly degrades the diversity of the subset, thus obscuring the effect of higher divergence. We therefore employ a balanced-sampling algorithm introduced by Jung et al. \cite{prismatic-synthesis} to match the diversity of all subsets in the embedding space\footnote{We stop at $N = 10k$ primarily due to data availability: after filtering the existing mixture for delta prompts (which removes $\sim$70\% of the pool) and then matching diversity with balanced-sampling, the largest subset we can sample from the 139k pool is on the order of 10k.}. Finally, we run on-policy distillation on each subset using Qwen3-VL-235B-Thinking as the teacher and Qwen3-VL-8B-Thinking as the student, and average the performance across 4 standard benchmarks---CharXiv, InfoVQA, ChartQAPro, SEEDBench2-Plus \cite{charxiv, infovqa, chartqapro, sb2p}. For further experimental details, see \S \ref{app:experimental_details}.


\paragraph{Answer divergence is critical for effective scaling.} The results are shown in Fig. \ref{fig:scaling_and_diversity} (Left). We first observe that even with just 500 prompts from any subset, models quickly improve by $\sim 1\%$ in average performance---we posit that this phenomenon is likely a baseline improvement, enabled by the exposure to dense token feedback from a strong teacher regardless of the nature of prompts. But after this point, the three subsets show strikingly different scaling---\textit{zero-delta prompts} ($\Delta = 0$) quickly saturate, with stagnating performance even when using 20 times more prompts than the baseline point. With \textit{delta prompts} ($\Delta > 0$), on the contrary, performance scales log-linearly with dataset size, showing consistent improvement with more prompts. Overall, these results demonstrate that answer divergence is critical for reliable scaling of distillation: \textit{delta prompts} drive continued learning, while \textit{zero-delta prompts} exhaust their value almost immediately.

\subsection{Do off-the-shelf datasets provide enough divergence?} \label{sec:do_off_the_shelf_datasets_provide_enough_divergence}
\begin{figure*}[t]
    \centering
    \includegraphics[width=.96\textwidth]{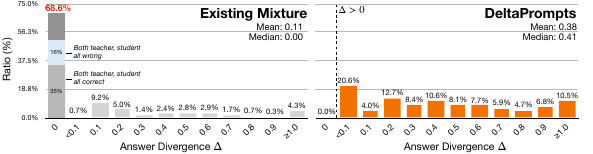}
    \caption{(Left) Divergence distribution of an off-the-shelf chart / document reasoning data mixture. \textbf{69\% of the existing prompts are zero-delta---\ie the teacher and student generate the exact same answer distribution---producing minimal learning value.} (Right) Divergence distribution of our synthesized dataset, \dataset. \textbf{Our pipeline not only removes all zero-delta prompts, but also produces a well-dispersed distribution of non-zero divergence.}}
    \vspace{-5pt}
    \label{fig:divergence_distributions}
\end{figure*}

\paragraph{Majority of prompts in existing datasets are zero-delta.} We additionally investigate whether existing datasets offer enough divergence for distillation. Using the same datasets and teacher-student pair as in \S \ref{sec:impact_of_answer_divergence}, we plot the answer divergence of each prompt in a quantized distribution in Fig. \ref{fig:divergence_distributions} and \S \ref{app:answer_divergence_distribution}. Here, we unexpectedly find that the majority of problems in existing datasets are effectively zero-delta---the two models produce homogeneous answer distribution across 16 rollouts, for up to 69\% of the problems. The result explains the random subset's scaling behavior in Fig. \ref{fig:scaling_and_diversity} (Left): as majority of samples in existing datasets are indeed zero delta, a random subset will highly likely inherit this property, leading to saturated performance similar to zero-delta subset.

\paragraph{Divergence is often at odds with diversity.} The above observation motivates us to ask a natural question---can't we just use the remaining 31\% of the prompts for distillation? Here, we find that naively filtering zero-delta prompts is suboptimal, as the process leads to significant degradation in prompt diversity. In Fig. \ref{fig:scaling_and_diversity} (Right), we compare the diversity of the distinct subsets with E-Vendi \cite{prismatic-synthesis}, a diversity metric known to better correlate with the trained model's generalization\footnote{Following Jung et al. \cite{prismatic-synthesis}, we use gte-Qwen-7B-Instruct \cite{gte-embedding} as the embedding model for E-Vendi.}. Across all datasets, we find that delta prompts exhibit significantly lower E-Vendi than random or zero-delta subsets, implying that the divergence in these datasets stems from a narrower set of systematic failure modes rather than a broad range of capability gaps.


\section{Synthesizing \dataset} \label{sec:synthesizing_deltaprompts}


Our analyses in \S \ref{sec:which_prompts_are_fit} motivate an alternative use of existing data---rather than training on the prompts directly, we repurpose them as seeds to synthesize better ones. In principle, one might hope to sidestep this issue by simply scaling the seed pool: if 31\% of prompts are delta prompts, a larger dataset would yield a larger filtered subset. In practice, however, high-quality multimodal reasoning data remains scarce \cite{bee, vero}, and more fundamentally, filtering alone cannot resolve the diversity-divergence tension we identify in \S \ref{sec:do_off_the_shelf_datasets_provide_enough_divergence}---the filtered subset is always a narrower slice of the original distribution. Below, we introduce two scalable methods that repurpose existing data as seeds, both leveraging the teacher model itself to generate divergence-inducing prompts at scale.

\subsection{Synthesis Pipeline}
\paragraph{Seed-guided Generation} As the most straightforward approach, we generate novel problems by prompting the teacher with existing prompts as few-shot examples. We first sample a set of $N$ image-prompt pairs $(x_i, I_i)_{i=1}^{N}$ from a seed dataset. Given a new image $I'$ (either drawn from the same dataset or a separate image source), we prompt the model to create a novel problem $x'$ for $I'$:
\begin{equation}
    x' \sim \pi_T\left(\cdot|(x_1, I_1), \cdots, (x_N, I_N), I'\right)
\end{equation}
To ensure the generated problems are well-defined, we additionally apply consistency-based filtering: for each generated $x'$, we sample $K$ independent responses from the teacher and retain only those where the teacher produces a consistent answer across majority of responses. This allows us to remove ill-posed or ambiguous problems from our pool that cannot even be reliably solved by the data generator. Seed-guided generation on its own does not explicitly target higher divergence; we subsequently employ rejection stage below to filter out zero-delta candidates.

\paragraph{Skill-guided Generation} Despite the scalability of seed-guided generation, it does not explicitly search for the types of problems that teacher and student diverge. Inspired by earlier works on skill-guided data curation \cite{metacognitive-capabilities, instruct-skillmix, instag}, we introduce a complementary skill-guided generation that strategically samples more from the student's failure modes.

The process proceeds in two steps. We first perform skill extraction: for each prompt $x$ with non-zero $\Delta$ in our candidate pool, we sample a corresponding student trajectory $\tau_\theta$ and a teacher trajectory $\tau_{T}$ such that the two trajectories disagree in final answers, \ie $\mathcal{E}(\tau_\theta) \neq \mathcal{E}(\tau_T)$. Skills are then extracted by contrasting these two trajectories---we prompt the teacher to identify the reasoning skill present in $\tau_T$ but absent in $\tau_\theta$, in a form of short natural language phrase (\eg \textit{interpolating logarithmic scales in a chart}). Iterating the extraction over all prompts $x$ in our pool, we obtain a set of discriminative skills that were missing in the student but were exhibited by the teacher. Finally, we utilize this skill pool by further conditioning the data generator on $K$ sampled skills:
\begin{equation}
    x' \sim \pi_T\left(\cdot|(x_1, I_1), \cdots, (x_N, I_N), s_1, \cdots, s_K, I'\right)
\end{equation}
The same consistency-based filtering is applied to ensure solvability of the generated prompts. For few-shot demonstrations, we use the synthesized prompts from the seed-guided generation---naturally forming a two-stage pipeline---and analyze the impact of this choice in \S \ref{sec:seed_guided_vs_skill_guided}.

\paragraph{Rejection for Answer Divergence} Finally, we measure the answer divergence $\Delta(x)$ for each generated prompt and reject all zero-delta candidates. Viewed broadly, the role of synthesis in our pipeline is to produce an abundance of diverse candidates, so that we can afford to be selective—retaining only the prompts with non-zero divergence. In \S \ref{sec:data_quality_analysis}, we show that this framework not only yields a substantially higher fraction of delta prompts than filtering existing datasets (57\% vs. 31\%), but also retains higher diversity after filtering.

\subsection{\dataset}
Using our pipeline, we generate \dataset, a dataset of 200k divergence-inducing prompts for chart \& document reasoning and high-resolution, perception-centric reasoning. We leverage state-of-the-art VLMs---Qwen3-VL-8B-Thinking and Qwen3-VL-235B-Thinking---as the target model pair for distillation.

We set the number of few-shot examples $N=1$ and number of sampled skills $K=10$, based on our preliminary observation that the number of few-shot examples has minimal impact on generation quality, while more sampled skills improve prompt diversity. We utilize publicly available data sources as both our seed prompts and image sources, and run both seed-guided and skill-guided generation, producing 110k and 90k prompts respectively in our two-stage process. We provide further details on our synthesis pipeline including the specific prompts and data sources in \S \ref{app:generating_deltaprompts}.

\section{Experimental Results}\label{sec:experimental_results}
\subsection{Data Quality Analysis} \label{sec:data_quality_analysis}
\subsubsection{Does \dataset improve beyond the seed data it originated from?} 
\begin{wrapfigure}{r}{.54\textwidth}
    \centering
    \renewcommand{\arraystretch}{1.0}
    \small
    \vspace{-10pt}
    \resizebox{.54\textwidth}{!}{
    \begin{tabular}{ lcccc }\toprule
          \multirow{2}{*}{\textbf{Dataset}} & \multicolumn{2}{c}{\textbf{Divergence}} & \multicolumn{2}{c}{\textbf{E-Vendi}} \\\cmidrule(lr){2-3}\cmidrule(lr){4-5}
          & \textbf{Mean (Median)} & \textbf{Var.} & \textbf{Pre-filter} & \textbf{Post-filter} \\
          \midrule
          Existing Mixture & 0.11 (0.00) & 0.063 & 141.67 & 108.19 \\
          \dataset & 0.38 (0.41) & 0.095 & 162.46 & 136.38 \\
          \bottomrule
    \end{tabular}
    }
    \captionof{table}{Comparison of data quality between existing mixture and \dataset. \textbf{\dataset shows significantly higher divergence and diversity than the existing mixture it originated from.}}
    \vspace{-5pt}
    \label{tab:delta_prompt_vs_existing_mixture}
\end{wrapfigure}

Does our synthesis pipeline actually improve the data quality beyond the seed datasets it starts from? In Figure \ref{fig:divergence_distributions} and Table \ref{tab:delta_prompt_vs_existing_mixture}, we validate this by comparing \dataset against existing datasets, in terms of their answer divergence and diversity. Compared to the existing datasets in Figure \ref{fig:divergence_distribution_baselines}, \dataset substantially improves answer divergence---our dataset is not only clear of any zero-delta prompts (as expected from the rejection protocol), but its distribution is also far more dispersed across the non-negative range. In addition, we observe consistent improvements in diversity as measured by E-Vendi, both before and after filtering for non-zero answer divergence (Table \ref{tab:delta_prompt_vs_existing_mixture}). Overall, the result confirms that our synthesis approach improves divergence and diversity beyond the seed data, rather than producing a homogeneous augmentation to the existing prompts.

\subsubsection{Understanding Seed-guided vs. Skill-guided Generation} \label{sec:seed_guided_vs_skill_guided}
\begin{wrapfigure}{r}{.52\textwidth}
    \centering
    \renewcommand{\arraystretch}{1.0}
    \small
    \vspace{-10pt}
    \resizebox{.52\textwidth}{!}{
    \begin{tabular}{ lcc }\toprule
          \textbf{Synthesis Pipeline} & \textbf{Yield} & \textbf{E-Vendi} \\
          \midrule
          Sampling from Existing Mixture & 0.31 & 108.19 \\
          Seed-guided Generation & 0.44 & 128.64 \\
          \parbox{6.3cm}{Skill-guided Generation w/ seed-guided prompts} & 0.57 & 142.47 \\
          \parbox{6.3cm}{Skill-guided Generation w/ off-the-shelf prompts} & 0.46 & 127.33 \\
          \bottomrule
    \end{tabular}
    }
    \captionof{table}{Comparison of synthesis pipelines. \textbf{Skill-guided generation performs best}, with seed-guided outputs as its most effective starting point.}
    \vspace{-5pt}
    \label{tab:seed_guided_vs_skill_guided}
\end{wrapfigure}
We also compare the effectiveness of the two stages in our pipeline---seed-guided generation and skill-guided generation. We specifically measure two metrics: the yield of delta prompts (\ie the fraction of prompts retained after rejection) and the diversity of those filtered prompts. 

In Table \ref{tab:seed_guided_vs_skill_guided}, we find that skill-guided generation leads to better yield and higher diversity compared to seed-guided generation, aligning with our intuition that the extracted skills can better capture the capability gap between the two models. However, when we directly run the skill-guided generation with off-the-shelf prompts as seeds (\ie bypassing Stage 1), both metrics fall back to the level of seed-guided generation. This suggests that the superiority of our second stage builds on the expanded data pool from seed-guided generation, and therefore, the two stages complement each other rather than one method strictly subsuming another.

\subsection{On-Policy Distillation with \dataset}
\subsubsection{Experimental Setup} \label{sec:experimental_setup}
Next, we validate \dataset by running on-policy distillation with our target teacher-student pair (Qwen3-VL-235B-Thinking and Qwen3-VL-8B-Thinking). We implement the algorithm using verl \cite{verl}, and use vllm \cite{vLLM} to run an external server that returns teacher model's log probability given a sequence. We train the student for 1 epoch on \dataset, and learn a single unified model for both chart \& document reasoning and high-resolution perception-centric reasoning. We refer to this model as \textbf{Qwen3-8B-DeltaThinker}.

For evaluation, we use 10 widely-used benchmarks across the two domains, as shown in Table \ref{tab:opd_results}. We compare against both off-the-shelf reasoning models \cite{qwen3-vl, glm} and the strongest open-recipe models from the community \cite{ares, revisual-r1, vision-r1, bee}. Notably, we find that reported results from the baselines often depend on differing inference protocols (\eg the choice of LLM judge for answer verification \cite{vlmevalkit}) and incompatible versions of evaluation frameworks, making direct comparison with published numbers unreliable. For fair comparison, we rerun all models across all benchmarks for 5 independent runs and report the standard error. We refer readers to \S \ref{app:experimental_details} for further experimental details.

\begin{table*}[t]\centering
\vspace{-15pt}
\resizebox{.99\textwidth}{!}{
    \begin{tabular}{ lccccc }\toprule
        \textbf{Model} & \textbf{CharXiv$_\text{RQ}$} & \textbf{InfoVQA$_\text{val}$} & \textbf{ChartQAPro$_\text{clean}$} & \textbf{SEEDBench2-Plus} & \textbf{Average} \\
        \midrule
        Qwen3-VL-8B-Thinking \cite{qwen3-vl} & 53.14 (0.07) & 85.33 (0.12) & 63.33 (0.11) & 69.87 (0.10) & 67.92 (0.04) \\
        GLM-4.1V-9B-Thinking \cite{glm} & 53.76 (0.06) & 85.04 (0.09) & 66.04 (0.13) & 73.03 (0.10) & 69.47 (0.04) \\
        \midrule
        Bee-8B-RL \cite{bee} & 57.00 (0.08) & 69.26 (0.10) & 65.96 (0.11) & 68.55 (0.11) & 65.19 (0.06) \\
        REVisual-R1 \cite{revisual-r1} & 41.50 (0.07) & 80.01 (0.13) & 59.93 (0.12) & 70.14 (0.12) & 62.90 (0.07) \\
        Vision-R1-7B \cite{vision-r1} & 37.28 (0.07) & 78.79 (0.09) & 44.14 (0.10) & 69.57 (0.14) & 57.45 (0.05) \\
        ARES-RL-7B \cite{ares} & 43.22 (0.08) & 81.29 (0.07) & 63.93 (0.15) & 68.77 (0.12) & 64.30 (0.06) \\
        \midrule
        \textbf{Qwen-8B-DeltaThinker} & \textbf{60.52 (0.09)} & \textbf{86.53 (0.11)} & \ul{71.09 (0.09)} & \ul{73.43 (0.11)} & \ul{72.89 (0.05)} \\
        \textbf{GLM-9B-DeltaThinker} & \ul{59.10 (0.10)} & \ul{85.78 (0.09)} & \textbf{72.55 (0.12)} & \textbf{74.81 (0.13)} & \textbf{73.06 (0.07)} \\
        \bottomrule \\
    \end{tabular}
}
\resizebox{.99\textwidth}{!}{
    \begin{tabular}{ lccccccc }\toprule
        \multirow{2}{*}[-2pt]{\textbf{Model}} & \multirow{2}{*}[-2pt]{\textbf{RealWorldQA}} & \multirow{2}{*}[-2pt]{\textbf{HRBench$_\text{4K}$}} & \multirow{2}{*}[-2pt]{\textbf{V*}} & \multicolumn{3}{c}{\textbf{VisualProbe}} & \multirow{2}{*}[-2pt]{\textbf{Average}} \\\cmidrule{5-7}
        & & & & \textbf{Easy} & \textbf{Medium} & \textbf{Hard} & \\
        \midrule
        Qwen3-VL-8B-Thinking & 73.07 (0.08) & 72.22 (0.05) & 76.96 (0.16) & 34.75 (0.23) & 13.61 (0.17) & 10.86 (0.44) & 46.92 (0.15) \\
        GLM-4.1V-9B-Thinking & 72.19 (0.08) & 73.75 (0.05) & 79.58 (0.19) & 39.01 (0.27) & 17.91 (0.15) & 16.04 (0.39) & 49.75 (0.18) \\
        \midrule
        Bee-8B-RL & 72.03 (0.10) & 60.15 (0.09) & 58.12 (0.16) & 26.95 (0.23) & 14.93 (0.15) & 10.40 (0.48) & 40.43 (0.16) \\
        REVisual-R1 & 64.36 (0.12) & 71.88 (0.05) & 69.11 (0.16) & 43.97 (0.23) & \ul{20.15 (0.17)} & \ul{18.87 (0.44)} & 48.06 (0.19) \\
        Vision-R1-7B & 67.58 (0.10) & 75.38 (0.05) & 81.15 (0.16) & 36.03 (0.23) & 16.40 (0.15) & 15.47 (0.39) & 48.67 (0.13) \\
        ARES-RL-7B & 66.67 (0.08) & 72.13 (0.05) & 71.20 (0.19) & 38.94 (0.27) & 18.50 (0.15) & 11.24 (0.44) & 46.45 (0.17) \\
        \midrule
        \textbf{Qwen-8B-DeltaThinker} & \ul{75.82 (0.08)} & \ul{77.25 (0.09)} & \ul{82.73 (0.19)} & \textbf{51.77 (0.23)} & \ul{20.15 (0.15)} & 16.98 (0.44) & \ul{54.12 (0.15)} \\
        \textbf{GLM-9B-DeltaThinker} & \textbf{77.04 (0.10)} & \textbf{80.25 (0.09)} & \textbf{84.25 (0.16)} & \ul{50.04 (0.27)} & \textbf{21.55 (0.15)} & \textbf{20.04 (0.39)} & \textbf{55.53 (0.13)} \\
        \bottomrule
    \end{tabular}
}
\caption{On-policy distillation result on chart \& document reasoning (Top) and real-world perception-centric reasoning (Bottom). \textbf{\dataset enables up to 15\% relative improvement on average across benchmarks, for both the target teacher-student pair (Qwen-8B-DeltaThinker) and an entirely new model family (GLM-9B-DeltaThinker).} We perform 5 independent runs across all baselines and report the standard error.}
\vspace{-10pt}
\label{tab:opd_results}
\end{table*}

\subsubsection{Main Result}
The results in Table \ref{tab:opd_results} show that \textbf{Qwen3-8B-DeltaThinker significantly improves over Qwen3-VL-8B-Thinking---7\% relative improvement in chart \& document reasoning and 15\% in real-world perception-centric reasoning, averaged across benchmarks}. These improvements are particularly notable given that our base model, Qwen3-VL-8B-Thinking, is itself a highly optimized reasoning model that already outperforms all open-recipe baselines---yet DeltaThinker still yields substantial gains on the hardest benchmarks such as CharXiv and VisualProbe. Compared to the open-recipe models, which employ staged training pipelines and order-of-magnitude more training samples (\eg Bee-8B-RL aggregates 15M samples across 150 sources with two-stage SFT followed by RL), our approach achieves stronger results with significantly less data, demonstrating the sample efficiency of targeted prompt synthesis. These results suggest that distillation on a high-quality set of prompts targeted for the student could outweigh heuristic aggregation of existing datasets.

\subsubsection{\dataset Generalizes Better to Out-of-Domain Tasks}
Motivated by prior reports that OPD mitigates catastrophic forgetting in supervised fine-tuning \cite{self-distillation-continual-learning}, we examine whether DeltaThinker maintains the strong performance of the base model on out-of-domain tasks such as math, puzzle and STEM reasoning. 

The results on MathVista, LogicVista and MMMU-Pro \cite{mathvista, logicvista, mmmu-pro} in Table \ref{tab:ood_results} present an unexpected finding: \textbf{ our model not only maintains, but \textit{improves} over the base model across all three benchmarks.} Importantly, we find that such gains are not universal to all OPD runs---when running OPD with 10k subset of existing data, the model sees early degradation on math and puzzle, whereas \dataset, on the same scale, either maintains or improves performance. This demonstrates that out-of-domain generalization depends not only on the choice of learning algorithm, but equally on the divergence and diversity of prompts it trains on.

\subsubsection{Does Answer Divergence Really Matter?}
\begin{figure*}[t]
    \centering
    \includegraphics[width=.96\textwidth]{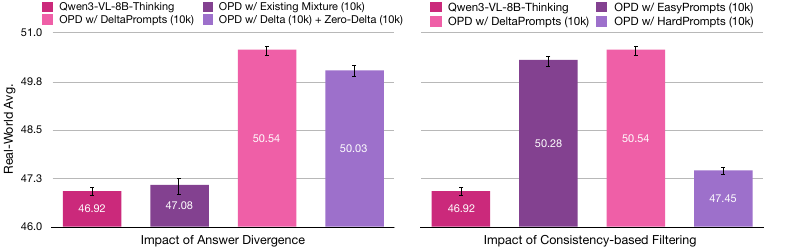}
    \caption{(Left) Ablation on answer divergence. \textbf{\dataset leads to the best results even under a data size-controlled setup, while adding additional zero-delta data does not help.} (Right) Ablation on consistency-based filtering. \textbf{Further optimizing for teacher consistency (EasyPrompts) yields no gain, while training on prompts the teacher cannot reliably solve (HardPrompts) degrades performance.}}
    \label{fig:ablation_studies}
\end{figure*}
Despite the compelling performance of \dataset, it remains a question whether the gains are truly driven by improving divergence in the training prompts, or simply by distilling a stronger teacher on more data. We analyze this via a series of controlled training runs, where we run distillation with (1) a random-sampled 10k subset of existing mixture, (2) a subset of \dataset with the same size, and (3) a 20k subset comprising 10k samples from \dataset and 10k zero-delta prompts rejected in our pipeline. The result in Figure \ref{fig:ablation_studies} (Left) shows that even in this strictly controlled setup, OPD with \dataset shows clear improvement over both the base model and direct OPD on the existing mixture. Furthermore, we find that augmenting \dataset with zero-delta prompts does not help, but degrade the performance---further confirming that zero-delta prompts act as noise rather than useful signal, even when they double the training set size.

\subsubsection{Does Consistency-based Filtering Matter?}
During the synthesis of \dataset, we improve the solvability of generated prompts by sampling 16 teacher responses per prompt and retaining those with $\text{teacher avg@16} \ge 0.5$, \ie the teacher produces a consistent answer across majority of responses. We ablate the impact of this consistency-based filtering by defining two subsets of real-world perception-centric prompts: (1) \textit{EasyPrompts}---prompts with $\text{teacher avg@16} = 1$, \ie teacher produces a consistent answer across all 16 responses, and (2) \textit{HardPrompts}---prompts with $\text{teacher avg@16} < 0.5$, \ie even the teacher fails to produce consistent answer for majority of responses. As with \dataset, we ensure that all subsets have no zero-delta prompts, strictly isolating the effect of teacher consistency. Controlling for data scale (10k), we run distillation on each subset and compare average performance.

In Fig. \ref{fig:ablation_studies} (Right), we find that further optimizing for teacher consistency ($\text{avg@16} = 1$) does not help, with \textit{EasyPrompts} marginally underperforming \dataset. The result validates our choice of a permissive consistency threshold, which increases the yield of the pipeline without sacrificing downstream performance. In contrast, training on \textit{HardPrompts} degrades performance---aligning with our intuition that teacher feedback becomes noisier with overly hard or ill-defined prompts.

\subsection{Generalization to a Novel Teacher-Student Pair}
Our synthesis pipeline aims to expose the gap between a specific pair of teacher and student, raising a natural question: can we reuse \dataset for a new pair of models without re-running the full pipeline? We test this by distilling with an entirely different model family—GLM-4.5V as teacher and GLM-4.1V-9B-Thinking as student. Instead of regenerating prompts from scratch, we directly reuse \dataset, simply filtering for non-zero answer divergence under the new pair. This lightweight rejection step retains 154k of the original 200k prompts.

The results are presented in the bottom rows of Table \ref{tab:opd_results}. Consistent with our findings on Qwen3-VL, \dataset yields coherent improvements across all benchmarks for the new pair of models. Overall, the results imply that (1) our prompts capture generalizable capability gaps that transfer across model families, rather than overfitting to the idiosyncrasies of a specific teacher-student pair, and (2) the rejection step---filtering for non-zero divergence---is sufficient to adapt the dataset, making our pipeline practical beyond the original models it was designed for.

\subsection{Reusing \dataset for Off-Policy Fine-Tuning}
\begin{table*}[t]\centering
\resizebox{.99\textwidth}{!}{
    \begin{tabular}{ lccccc }\toprule
        \textbf{Model} & \textbf{Chart \& Document Avg.} & \textbf{Real-World Avg.} & \textbf{MathVista$_\text{testmini}$} & \textbf{LogicVista} & \textbf{MMLU-Pro$_\text{vision}$} \\
        \midrule
        Qwen3-VL-8B-Instruct & 66.85 (0.05) & 47.82 (0.13) & \ul{77.24 (0.25)} & 57.21 (0.29) & 42.14 (0.23) \\
        \quad + SFT on Existing Mixture & 67.52 (0.04) & 49.50 (0.18) & 74.50 (0.20) & 55.05 (0.23) & 41.10 (0.20) \\
        \quad + \textbf{SFT on \dataset} & \textbf{71.57 (0.05)} & \ul{55.04 (0.17)} & 75.20 (0.25) & \ul{57.27 (0.23)} & \textbf{47.74 (0.16)} \\
        \quad + \textbf{OPD on \dataset} & \ul{69.32 (0.06)} & \textbf{55.17 (0.15)} & \textbf{77.70 (0.24)} & \textbf{58.87 (0.19)} & \ul{46.34 (0.23)} \\
        \bottomrule \\
    \end{tabular}
}
\caption{Distilling a non-reasoning model (Qwen3-VL-8B-Instruct) using \dataset. \textbf{SFT on \dataset forgets less than the existing mixture on out-of-domain tasks, while staying competitive to on-policy distillation on in-domain tasks.} We generate 4 teacher trajectories per prompt without additional filtering.} 
\label{tab:sft_results}
\end{table*}
Beyond on-policy distillation, we investigate whether \dataset can serve as an off-policy resource for standard supervised fine-tuning (SFT). We generate 4 trajectories per prompt using Qwen3-VL-235B-Thinking and train a non-reasoning model, Qwen3-VL-8B-Instruct on this data via SFT. We did not perform any additional curation such as trajectory-level filtering, in order to simulate the most basic usage of this dataset focusing on prompt quality. We also run OPD with the same student model for direct comparison.

The results in Table \ref{tab:sft_results} show that for in-domain tasks (Chart \& Document Avg. and Real-World Avg.), SFT on \dataset substantially outperforms both the base model and SFT on existing mixture, matching or exceeding direct OPD on the non-reasoning model. On out-of-domain benchmarks, SFT on \dataset does show degraded performance compared to OPD, consistent with prior observations that SFT is prone to forgetting \cite{sft-memorizes, why-does-rl-generalize-better-than-sft}. However, when compared to SFT on existing mixture, \dataset notably generalizes better to out-of-domain tasks while still delivering stronger in-domain performance---suggesting that the quality of the training prompts partially compensates for the algorithmic limitations of SFT.
\section{Conclusion} \label{sec:conclusion}
We identify a critical data bottleneck in multimodal distillation: the prevalence of zero-delta prompts in standard datasets that impede effective learning of the student model. We establish \textit{answer divergence} as a principled metric for quantifying the capability gap between the teacher and student. To systematically source high-divergence prompts, we propose a two-stage synthesis pipeline that performs targeted rejection sampling for the student's failure modes. Our resulting dataset, \dataset, yields substantial and highly transferable gains across diverse reasoning benchmarks, and generalizes better to out-of-domain tasks, even when distilling into highly optimized student models.

\section*{Acknowledgments}
This work was supported by the Institute of Information \& Communications Technology Planning \& Evaluation (IITP) grant (No. RS-2024-00457882, National AI Research Lab Project).

\bibliography{neurips_2026}
\bibliographystyle{abbrv}

\newpage


\appendix
\section{Experimental Details} \label{app:experimental_details}
\subsection{Answer Divergence}
In our main experiments, we compute answer divergence using Qwen3-VL-235B-Thinking as the teacher and Qwen3-VL-8B-Thinking as the student. For each prompt, we sample 16 responses from both models, instructing each to end with \texttt{Final Answer: [ANSWER]} for easy rule-based extraction. We then use Qwen3-32B \cite{qwen3} as an LLM judge to group semantically equivalent answers, yielding empirical answer distributions $\widehat{P}_T(a|x)$ and $\widehat{P}_\theta(a|x)$. The answer divergence $\widehat{\Delta}$ is computed as the reverse KL between these distributions. When the supports do not fully overlap (\eg the student never produces an answer that the teacher generates), we apply Laplace smoothing—adding a pseudocount of 1 to all answer categories before normalization—to ensure no answer receives zero probability.

To analyze the impact of answer divergence (\S \ref{sec:impact_of_answer_divergence}), we create 3 distinct subsets---\textit{delta prompts}, \textit{zero-delta prompts}, and a random subset---from 4 existing datasets: ChartQA, SciVQA, InfoVQA, and arXivQA. We control for prompt diversity using the balanced-sampling approach of Jung et al. \cite{prismatic-synthesis} (presented in Algorithm \ref{alg:higher_diversity_sampling} for completeness). Specifically, we encode each prompt using an embedding model \cite{gte-embedding}, perform K-means clustering over the representations, then sample in a balanced manner---up-sampling from sparse clusters and down-sampling from dense ones. We tune the number of clusters to match the diversity of the most diverse subset (\textit{zero-delta prompts}). With this process, we reduce the diversity gap between subsets to within 1\%, compared to the 29.4\% degradation observed under naive sampling (Fig. \ref{fig:scaling_and_diversity}). For all diversity comparisons (Fig. \ref{fig:scaling_and_diversity}, Table \ref{tab:delta_prompt_vs_existing_mixture} and Table \ref{tab:seed_guided_vs_skill_guided}), we measure the embedding entropy of 10k subset random-sampled from each dataset to remove the effect of the dataset size.

\subsection{Generating \dataset} \label{app:generating_deltaprompts}
We use our teacher model (Qwen3-VL-235B-Thinking) as the data generator to produce \dataset. The specific prompts used for data generation and answer aggregation are provided in \S \ref{app:prompts}. For chart \& document reasoning, we use ChartQA \cite{chartqa}, InfoVQA \cite{infovqa}, arXivQA \cite{arxivqa}, SciVQA \cite{scivqa} as seed datasets, and for real-world perception-centric reasoning, we use DeepEyes47k \cite{deepeyes}, DriveAction \cite{drive-action} and VisualProbe$_\text{train}$ \cite{visualprobe}. We source images for the new prompts from arXivQA, SciVQA, PDF-VQA \cite{pdf-vqa} (chart \& document reasoning), and DeepEyes47k, VisualProbe$_\text{train}$, Div2k \& Flickr2k \cite{div2k-flickr2k}, and sama-drives-california \cite{sama-drives-california} (real-world perception-centric reasoning). We find that these image sources often contain near-duplicate images with minimal differences between frames; we deduplicate each source using perceptual hashing from ImageDedup \cite{imagededup}. 

To preempt data leakage, we additionally run the deduplication logic against all test images from our evaluation benchmarks (Table \ref{tab:opd_results}), filtering out any matches from our collected image pool. We define a contaminated image as one with a hamming distance of less than 10 (after hashing) from any test set image, following the default setup in ImageDedup \cite{imagededup}. Through this process, 3.1\% of collected images are flagged—all chart images from arXivQA or SciVQA—some of which are false positives detected due to visual similarity despite presenting different underlying data. We conservatively remove all flagged images from our synthesis pipeline before producing \dataset.

\subsection{Training with \dataset}
We implement on-policy distillation using verl \cite{verl}, an open-source framework for online RL with LLMs. We find that a naive implementation that colocates the student and teacher in a sequential training loop significantly degrades the throughput due to the scale of our teacher model (235B parameters). To address this, we run a separate vllm \cite{vLLM} server for the teacher's log-probability computation, enabling $\sim$512 parallel rollouts on a single GB200 node and hence accelerated training. We use 4 GB200 nodes for our training runs.

One practical consideration is that the log probabilities from the same model on the same sequence can differ between HuggingFace Transformers and vllm due to implementation differences. While this is a known issue in the community and the gap is small for our base models (\eg the average log-probability difference across 1000 test sequences was 0.04 for Qwen3-VL-8B-Thinking), it directly affects the training signal and future implementations should account for this discrepancy. Throughout our OPD experiments, we set learning rate = 5e-7, batch size = 128, and generate 4 rollouts per prompt. We train for 1 epoch with a maximum response length of 16384. In our SFT experiment, we use learning rate = 5e-5, batch size = 512, and train for 5 epochs.

\subsection{Evaluation}
We evaluate on 10 widely-used benchmarks across both domains: CharXiv \cite{chartqa}, InfoVQA \cite{infovqa}, ChartQAPro \cite{chartqapro}, SEEDBench2-Plus \cite{sb2p}, RealWorldQA \cite{rwqa}, HRBench \cite{hrbench}, V* \cite{v*}, and VisualProbe Easy / Medium / Hard \cite{visualprobe}. We use VLMEvalKit \cite{vlmevalkit} as our evaluation framework, and implement separate evaluation scripts for ChartQAPro and VisualProbe, which are not included in VLMEvalKit.

Throughout the evaluation, we follow VLMEvalKit's protocol: (1) generate outputs from the evaluated model using vllm, (2) verify model-generated answers against the ground truth via an LLM judge, and (3) aggregate accuracy. We use GPT-4o-mini \cite{4o-mini} as the judge for all evaluations, following VLMEvalKit's official recommendation. We make two dataset-specific modifications. For InfoVQA, the default metric in VLMEvalKit relies on naive string-overlap scoring, which leads to high false-positive and false-negative rates. Following Zhang et al. \cite{bee}, we replace the default metric with judge-based accuracy. For ChartQAPro, we find that a small portion of problems are structured as lists of multiple questions, making answer parsing in VLMEvalKit protocol unreliable. We evaluate only single-question problems, referring to this subset as ChartQAPro$_\text{clean}$.

Across all baselines, we use the recommended decoding parameters for each model, and default to Qwen3-VL's setup if not available. For Qwen-8B-DeltaThinker and GLM-9B-DeltaThinker, we use temperature = 0.7, top\_p = 0.8, top\_k = 20, and max output length = 32768.

 \begin{algorithm}[t]
\caption{\textit{Higher Diversity Sampling}}
\begin{algorithmic}
\Require Data representation $D \in \mathbb{R}^{|\mathcal{D}| \times d}$, number of clusters $k$ and target subset size $N_{\text{target}}$
\Ensure Indices of selected subset $S \subseteq \{1, \cdots, |\mathcal{D}|\}$
\State $S \gets \emptyset$ \Comment{Initialize the subset.}
\State $\{c_1, \cdots, c_{k}\} = \kMeans(D)$ \Comment{Cluster data. $c_i$ is a set of indices corresponding to cluster $i$.}
\While{$|S| < N_{\text{target}}$}
\State $c \gets \randomSample(\{c_1, \cdots, c_k\}, 1)$ \Comment{Randomly pick a sampling cluster.}
\State $S_{\text{new}} \gets \randomSample\left(c, \left\lceil \frac{N_{\text{target}}}{k} \right\rceil\right)$ \Comment{Sample new samples from the chosen cluster.}
\State $S \gets S \cup S_{\text{new}}$  \Comment{Add new samples to the subset.}
\EndWhile
\Return $S$ \Comment{Return the sampled subset.}
\end{algorithmic}
\label{alg:higher_diversity_sampling}
\end{algorithm}

\newpage

\section{Prompts} \label{app:prompts}
\begin{tcolorbox}[colback=white,colframe=black!75!white,colbacktitle=black!75!white,title=Prompt for Seed-guided Generation (Chart \& Document Reasoning)]
A problem for the first image:\\
\{reference\_question\}\\

Above are two images. For the first image, I have given you a reasoning problem that asks about the image. Your task is to create a harder problem for the second image.\\
- Do not just copy the problem verbatim from the example, and try to be creative.\\
- The new problem should be about the second image, and it should be similar in complexity or harder than the given problem.\\
- Make sure your problem leads to an objective answer without any subjectivity.\\
- It would be great if you could create a more natural problem that incorporates various reasoning skills in a meaningful way, rather than artificially asking for them.\\
- Try to solve your problem to make sure that it is not ill-defined, ambiguous, or unsolvable.\\
- Do not specifically refer to the second image as \textit{second image} in your question---the students will only be given one image. Just call it an \textit{image}.\\
- Do not leak the problem-relevant information in your question statement, when it can be inferred from the image.\\
- Answer in English.\\

Only output the final hard problem of yours and nothing else. Your output should be formatted as: \textit{Final Problem: your hard problem}.
\end{tcolorbox}

\begin{tcolorbox}[colback=white,colframe=black!75!white,colbacktitle=black!75!white,title=Prompt for Seed-guided Generation (Real-world Perception-centric Reasoning)]
A problem for the first image:\\
\{reference\_question\}\\

Above are two images. For the first image, I have given you a reasoning problem that asks about the image. Your task is to create a harder problem for the second image.\\
- Do not just copy the problem verbatim from the example, and try to be creative.\\
- The new problem should be about the second image, and it should be similar in complexity or harder than the given problem.\\
- Make sure your problem leads to an objective answer without any subjectivity.\\
- It would be great if you could create a more natural problem that incorporates various reasoning skills in a meaningful way, rather than artificially asking for them. You are encouraged to focus on hard-to-find small details in the second image when creating the problem.\\
- Try to solve your problem to make sure that it is not ill-defined, ambiguous, or unsolvable.\\
- Do not specifically refer to the second image as \textit{second image} in your question---the students will only be given one image. Just call it an \textit{image}.\\
- Do not leak the problem-relevant information in your question statement, when it can be inferred from the image.\\
- Answer in English.\\

Only output the final hard problem of yours and nothing else. Your output should be formatted as: \textit{Final Problem: your hard problem}.
\end{tcolorbox}

\newpage

\begin{tcolorbox}[colback=white,colframe=black!75!white,colbacktitle=black!75!white,title=Prompt for Skill Extraction]
You are an expert AI researcher specializing in error analysis for Multimodal Large Language Models. Your task is to analyze specific failure modes in visual reasoning tasks by comparing \textit{teacher trajectories} (reasoning that leads to the correct answer) against \textit{student trajectories} (reasoning that leads to an incorrect answer).\\

\#\#\# Analysis Framework\\
For the provided input, perform the following discriminative analysis:\\

1. \textit{Compare Visual Attention}: Where did the teacher look versus the student? Did the student miss a region entirely, or did they misinterpret what they saw? \\
2. \textit{Compare Logic/Semantics}: Did the student fail in reasoning despite seeing correctly? Check for arithmetic errors (\eg adding numbers wrong), misunderstanding definitions (\eg what counts as a ``car''), wrong logical deductions (\eg false implication, contradictions, or invalid spatial reasoning), or other logical fallacies. \\
3. \textit{Compare Root Cause}: Identify the core difficulty (\eg occlusion, visual camouflage, saliency bias, arithmetic failure, logical fallacy). \\

And then, finally, \\
4. \textit{Extract Generalizable Skill}: Generate a single phrase describing the abstract skill required based on the root cause and the core difficulty factor (\eg changing ``missed the car behind the van'' to ``instance segmentation failure due to occlusion'').\\

\#\#\# Final Answer\\
Please output the generalizable skill in a single phrase. Do not include any other text or comments.\\

\#\#\# Input Data\\

Question:\\
\{question\}\\

Teacher Trajectory (Correct):\\
\{teacher\_trajectory ($\tau_T$)\}\\

Student Trajectory (Incorrect):\\
\{student\_trajectory ($\tau_\theta$)\}
\end{tcolorbox}

\begin{tcolorbox}[colback=white,colframe=black!75!white,colbacktitle=black!75!white,title=Prompt for Skill-guided Generation (Both domains)]
A problem for the first image:\\
\{reference\_question\}\\

Above are two images. For the first image, I have given you a reasoning problem that asks about the image. Your task is to create a harder problem for the second image.\\

\#\#\# Target Reasoning Skills\\
Your problem should require one or more of the following reasoning skills:\\
- \{skill 1\}\\
- \{skill 2\}\\
...\\
- \{skill 10\}\\

\#\#\# Requirements\\
- Do not just copy the problem verbatim from the example, and try to be creative.\\
- The new problem should be about the second image, and it should be similar in complexity or harder than the given problem.\\
- Make sure your problem leads to an objective answer without any subjectivity.\\
- It would be great if you could create a more natural problem that incorporates various reasoning skills in a meaningful way, rather than artificially asking for them.\\
- Try to solve your problem to make sure that it is not ill-defined, ambiguous, or unsolvable.\\
- Do not specifically refer to the second image as \textit{second image} in your question---the students will only be given one image. Just call it an \textit{image}.\\
- Do not leak the problem-relevant information in your question statement, when it can be inferred from the image.\\
- Answer in English.\\

Only output the final hard problem of yours and nothing else. Your output should be formatted as: \textit{Final Problem: your hard problem}.
\end{tcolorbox}

\newpage

\begin{tcolorbox}[colback=white,colframe=black!75!white,colbacktitle=black!75!white,title=LLM Judge-based Answer Grouping]
Given a question and multiple student's answers to the question, group them into the groups of equivalent answers. The answers might be different in their surface forms but semantically equivalent. Do not attempt to solve the question or verify the correctness of the answers, just group them based on semantic equivalence. Your response should be in a json format where key is the answer (string) and the value is the list of students (string), formatted as:\\
\{\{\\
\hspace*{1.5em}``answer 1": \text{[student generation indices corresponding to answer 1]},\\
\hspace*{1.5em}``answer 2": \text{[student generation indices corresponding to answer 2]},\\
\hspace*{1.5em}...\\
\}\}\\

Example\\
Question:\\
On the lecture slide, first count the number of lines of text below the lecture title. Then, count the number of words in the course name ``Mathematics for Computer Science''. Next, count the number of letters in the instructor’s name. Finally, count the number of distinct colors in the gradient bar. Multiply these four values together. What is the result?\\

Student Answers:\\
Student 1: 512\\
Student 2: 484\\
Student 3: The final result is 484.\\
Student 4: 512\\
Student 5: None\\
Student 6: 29\\
Student 7: **Final Answer** 512.\\
Student 8: 484\\

Grouped Answers:\\
\{\{\\
\hspace*{1.5em}``512": [1, 4, 7],\\
\hspace*{1.5em}``484": [2, 5, 8],\\
\hspace*{1.5em}``29": [6],\\
\hspace*{1.5em}``None": [5],\\
\}\}\\

Question:\\
\{question\}\\

Student Answers:\\
\{answers\_str\}\\

Grouped Answers:
\end{tcolorbox}

\newpage

\section{Additional Results}
\subsection{Answer Divergence Distribution} \label{app:answer_divergence_distribution}
\begin{figure*}[h]
    \centering
    \includegraphics[width=.96\textwidth]{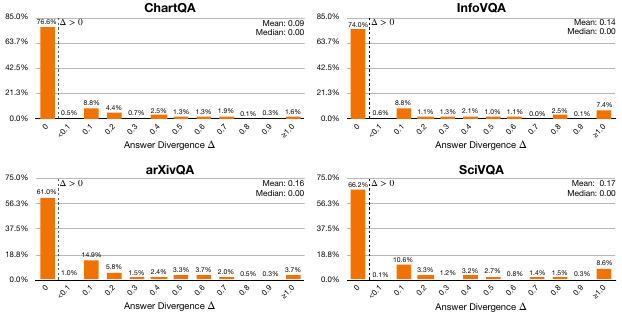}
    \caption{Distribution of answer divergence of the four datasets used in \S\ref{sec:do_off_the_shelf_datasets_provide_enough_divergence}, plotted individually. We use the same set of teacher-student models as in \S\ref{sec:do_off_the_shelf_datasets_provide_enough_divergence}. In all datasets, student and teacher tend to generate homogeneous output distributions, with zero-delta prompts comprising from 61\% up to 77\% of the entire distribution.}
    \vspace{-10pt}
    \label{fig:divergence_distribution_baselines}
\end{figure*}

\subsection{Out-of-Domain Generalization}
\begin{table*}[ht]
    \centering
    \resizebox{.75\textwidth}{!}{
    \begin{tabular}{ lccc }\toprule
          \textbf{Model} & \textbf{MathVista$_\text{mini}$} & \textbf{LogicVista} & \textbf{MMMU-Pro$_\text{Vision}$} \\
          \midrule
          Qwen3-VL-8B-Thinking & 81.12 & 56.64 & 53.41 \\
          OPD w/ Seed Data (10k) & 79.05 & 56.24 & 53.64 \\
          OPD w/ \dataset (10k) & 81.05 & 58.04 & 53.75 \\
          \midrule
          Qwen-8B-DeltaThinker & \textbf{81.74} & \textbf{61.97} & \textbf{56.30} \\
          \bottomrule
    \end{tabular}
    }
    \captionof{table}{Evaluation on Out-of-Domain Tasks. \textbf{Qwen-8B-DeltaThinker, despite specializing on chart / document / perception-centric reasoning, shows on-par or better performance than Qwen3-VL-8B-Thinking across math, puzzle and STEM reasoning.}}
    \label{tab:ood_results}
\end{table*}


\newpage
\section{Related Works} \label{app:related_works}
\paragraph{Algorithmic Advances in Distillation} The algorithmic foundations of knowledge distillation have been extensively refined, beginning with classical formulations that match teacher output distributions via forward KL \cite{hinton-kd, distillbert}. Subsequent work has proposed alternative divergence objectives to improve training stability and mitigate mode-seeking pathologies \cite{minillm, distillm, distillm-2, rethinking-kl, todi, f-divergence-minimization}, and has addressed the distribution shift between teacher and student through on-policy distillation \cite{gkd, opsd, opd-thinking-machines, opcd}. More recent variants extend this framework through self-distillation \cite{opsd, rl-via-self-distillation}, black-box settings where only teacher outputs are accessible \cite{lion, gad}, and auxiliary training signals \cite{entropy-aware-opd, self-distilled-rlvr}. Notably, Kim et al. \cite{self-distillation-degrades} observe that naive distillation can bias the student toward narrow patterns when the training data lacks coverage—similar in spirit to our finding that distilling on zero-delta prompts reinforces stagnation rather than driving learning. While these works have made substantial progress on the algorithmic front, they largely treat the training prompts as a given.

\paragraph{Data-Centric Perspectives on Distillation} Our work sits within a smaller but growing line of research that views distillation through a data-centric lens. Early evidence from Li et al. \cite{dynamic-kd} showed that adaptive data selection can match full-data performance with a fraction of the samples, and subsequent methods have introduced principled selection criteria based on student-teacher compatibility \cite{selective-reflective-distillation}, informative trajectory ranking \cite{which-reasoning-trajectories}, attention-head influence \cite{air}, or token-level acceptance rates \cite{selectkd, student-in-the-loop}. Most closely related to our work, Xu et al. \cite{tip} studies data importance in OPD, but focuses on token-level data selection for a fixed set of prompts. Complementing these efforts, DC-CoT \cite{dc-cot} provides a systematic benchmark of data-centric knobs for CoT distillation, while Jung et al. \cite{prismatic-synthesis} proposes gradient-based diversification for synthetic data generation. We differ in two key ways: our selection criterion is answer divergence---a principled teacher-student gap measure---and our pipeline actively synthesizes prompts targeted at this signal rather than filtering from a fixed pool.

\paragraph{Open Recipes for Multimodal Reasoning} In parallel, a rapidly growing body of work has released open recipes for post-training multimodal reasoning models \cite{internvl3, revisual-r1, vision-r1, bee, vero, mmfinereason, skyworkk-r1v3, r1-onevision}. These efforts typically aggregate existing datasets with heuristic quality filters and apply staged SFT or RL pipelines over teacher-generated trajectories. Lu et al. \cite{why-does-rl-generalize-better-than-sft} offers a direct data-centric analysis of this trend, arguing that RL's advantage over SFT stems from implicit filtering toward medium-difficulty samples, with Chu et al. \cite{sft-memorizes} reaching a similar conclusion through a broader comparison. Most relevant to our setting, Long Grounded Thoughts\cite{lgt} synthesizes visual problems at scale---though the work focuses on scaling reasoning-heavy trajectories rather than teacher-student divergence. Our work is complementary: we   isolate a principled quantity (answer divergence) as the guiding criterion, and show that prompt synthesis targeted at this signal yields substantial improvements across both on-policy and off-policy distillation regimes.

\newpage

\section{Limitations and Future Work}\label{app:limitations_and_future_work}
While \dataset establishes a robust foundation for divergence-driven distillation, our synthesis pipeline measures answer divergence against the teacher-student pair at the beginning of training, and uses the resulting dataset throughout. As the student evolves during distillation, the capability gap may shift---prompts that were once divergence-inducing may become zero-delta. We mitigate this in practice by training for a single epoch, so each prompt is seen only once, limiting the impact of drift. Still, a more active approach could iteratively re-measure divergence against the student's current state and regenerate targeted prompts. Additionally, our pipeline generates novel text prompts but still sources images from existing datasets rather than synthesizing them; a natural extension is to close this generative loop through diffusion-based generation or structured editing.

Beyond these methodological extensions, our empirical validation focuses on visual domains where images are abundant but high-quality reasoning data is sparse—chart, document, and real-world perception-centric reasoning. While these provide a clear testbed, the core principles of answer divergence are broadly applicable, and a particularly promising direction is extending this framework to long-horizon multimodal tasks such as autonomous web navigation, embodied agent control, or long-form video understanding. We view \dataset as a first step toward a more principled, data-centric approach to distillation.

\section{Broader Impacts}\label{app:broader_impacts}
This work demonstrates that carefully constructed prompt distributions can substantially improve the efficiency of knowledge distillation in vision-language models. On the positive side, our findings suggest a path toward training capable compact models with significantly fewer resources, potentially democratizing access to strong multimodal reasoning capabilities. The principled framework we introduce---measuring and optimizing for answer divergence---may also generalize beyond distillation to other settings where data quality matters more than quantity. On the other hand, more efficient distillation techniques could accelerate the proliferation of capable models with limited oversight, and the synthesis pipeline itself could be repurposed to generate misleading visual reasoning problems. We encourage practitioners to apply these methods responsibly and in accordance with established ethical guidelines for model development and deployment.

\newpage
\section{Qualitative Examples}

\subsection{Skill Extraction}
In Examples 1 to 4 below, we present qualitative examples of extracted skills from our synthesis pipeline, using the input problems generated via seed-guided generation. We include the key snippet from answer extraction trace generated by the teacher during skill extraction. Overall, the pipeline is able to discover precisely where the student trajectory diverged from the teacher's, and output these into a concise skill description---providing targeted signals for the subsequent generation stage to focus on the student's actual weaknesses.

\vspace{10pt}
\begin{tcolorbox}[colback=white,colframe=black!75!white,colbacktitle=black!75!white,title=Skill Extraction Example 1]
  \textbf{Input Problem}
      \begin{center}
      \includegraphics[width=0.45\linewidth]{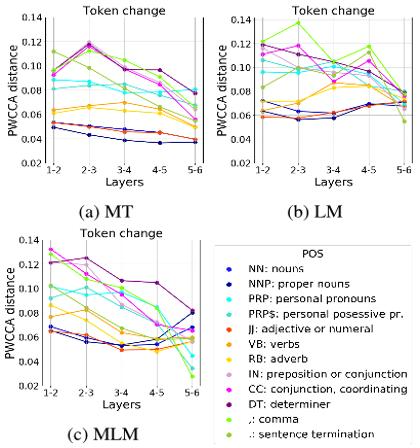}
    \end{center}
  How many part-of-speech tags in the second image exhibit a peak PMWCA distance within the 2-3 layer interval in all three subplots (a), (b) and (c), and additionally show a consistent decline in PMWCA distance from the 4-5 to 5-6 layer interval across the same subplots?
  
  \tcbline
  \textbf{Skill Extraction Trace} \\[5pt]
  ... The student incorrectly claimed all 11 POS tags meet these conditions. The student appears to have misinterpreted the plots, likely by assuming all lines peak at 2-3 and decline from 4-5 to 5-6 \texttt{...[omitted]}\\
  The generalizable skill needed is the ability to validate patterns systematically across multiple data representations without overgeneralizing.
  
  \tcbline
  \textbf{Extracted Skill} \\[5pt]
  \textit{\textbf{Cross-referential pattern validation across multiple data representations}}
\end{tcolorbox}

\newpage
\begin{tcolorbox}[colback=white,colframe=black!75!white,colbacktitle=black!75!white,title=Skill Extraction Example 2]
  \textbf{Input Problem}
  \begin{center}
  \includegraphics[width=0.45\linewidth]{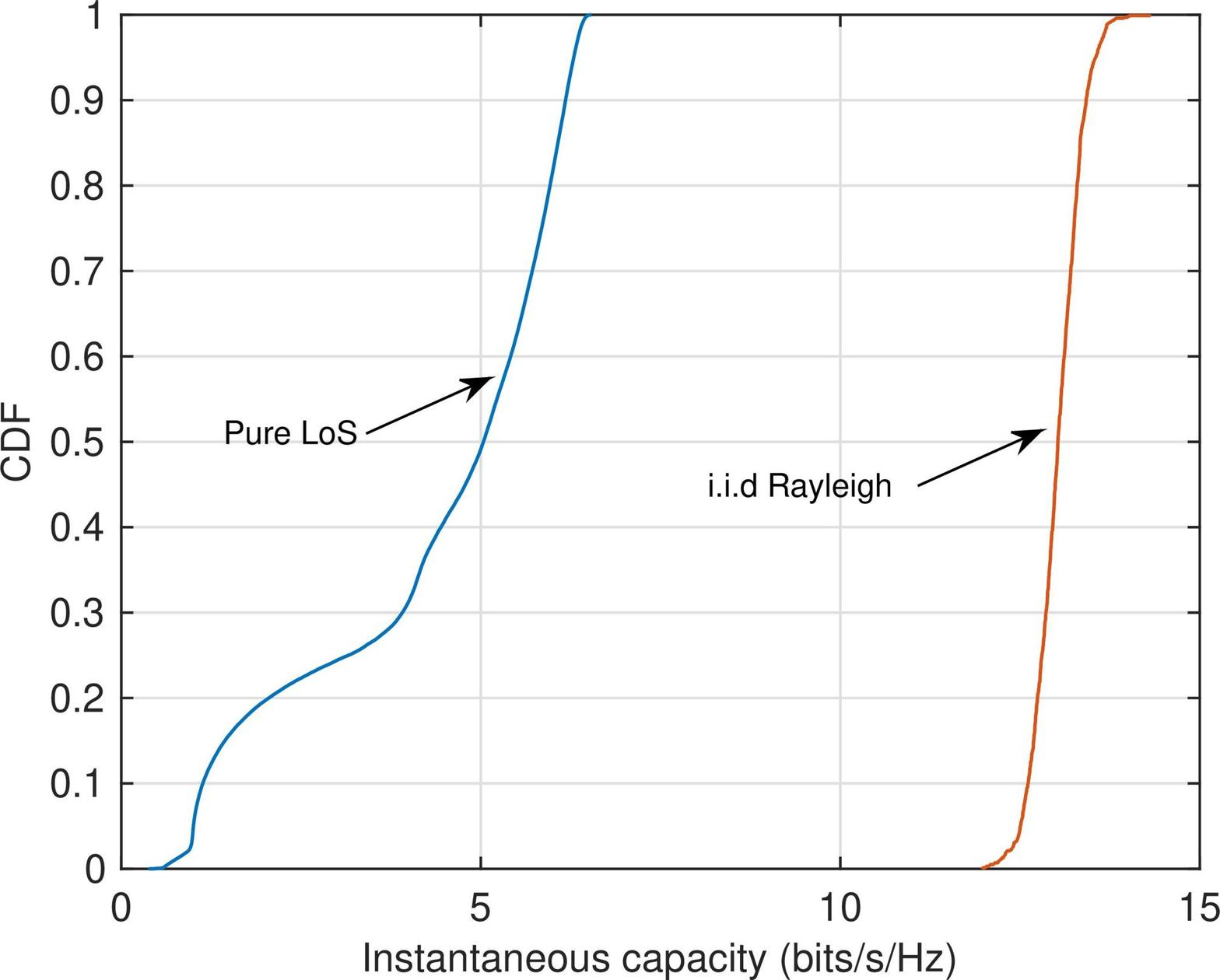}
  \end{center}
  For the 80\% probability mass range (between the 10th and 90th percentiles) of each distribution, calculate the width in bits/s/Hz. How many times wider is the Pure LoS distribution's range compared to the i.i.d Rayleigh distribution's range? Round your answer to the nearest whole number.
  
  \tcbline
  \textbf{Skill Extraction Trace} \\[5pt]
  ... Student misread the percentiles, interpreting the 10th percentile of Pure LoS as ~2.5 but  placing the 90th percentile at ~10.5 (should be ~6.5). Similarly, they misread the i.i.d Rayleigh range as 8.5–13.5 (should be 12.2–13.8) \texttt{...[omitted]}
  
  \tcbline
  \textbf{Extracted Skill} \\[5pt]
  \textit{\textbf{Spatial mapping of CDF values to x-axis quantiles}}
\end{tcolorbox}

\begin{tcolorbox}[colback=white,colframe=black!75!white,colbacktitle=black!75!white,title=Skill Extraction Example 3]
  \textbf{Input Problem}
  \begin{center}
  \includegraphics[width=0.45\linewidth]{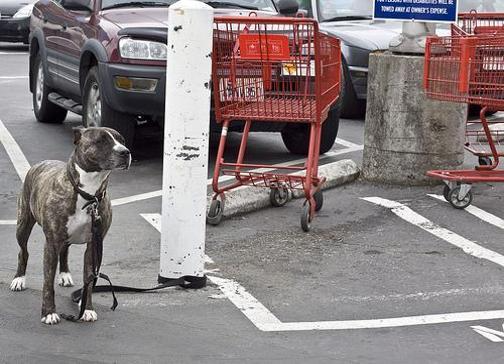}
  \end{center}
  What color is the text printed on the red shopping cart?
  
  \tcbline
  \textbf{Skill Extraction Trace} \\[5pt]
  ... Student thought ``Walmart carts have black text'', but the image might not be a Walmart cart, or the text color is different here. The student relied on a faulty assumption (saliency bias towards known brands) instead of visual evidence. \texttt{...[omitted]}
  
  \tcbline
  \textbf{Extracted Skill} \\[5pt]
  \textit{\textbf{Avoiding saliency bias from prior knowledge when interpreting visual evidence}}
\end{tcolorbox}

\begin{tcolorbox}[colback=white,colframe=black!75!white,colbacktitle=black!75!white,title=Skill Extraction Example 4]
  \textbf{Input Problem}
  \begin{center}
  \includegraphics[width=0.45\linewidth]{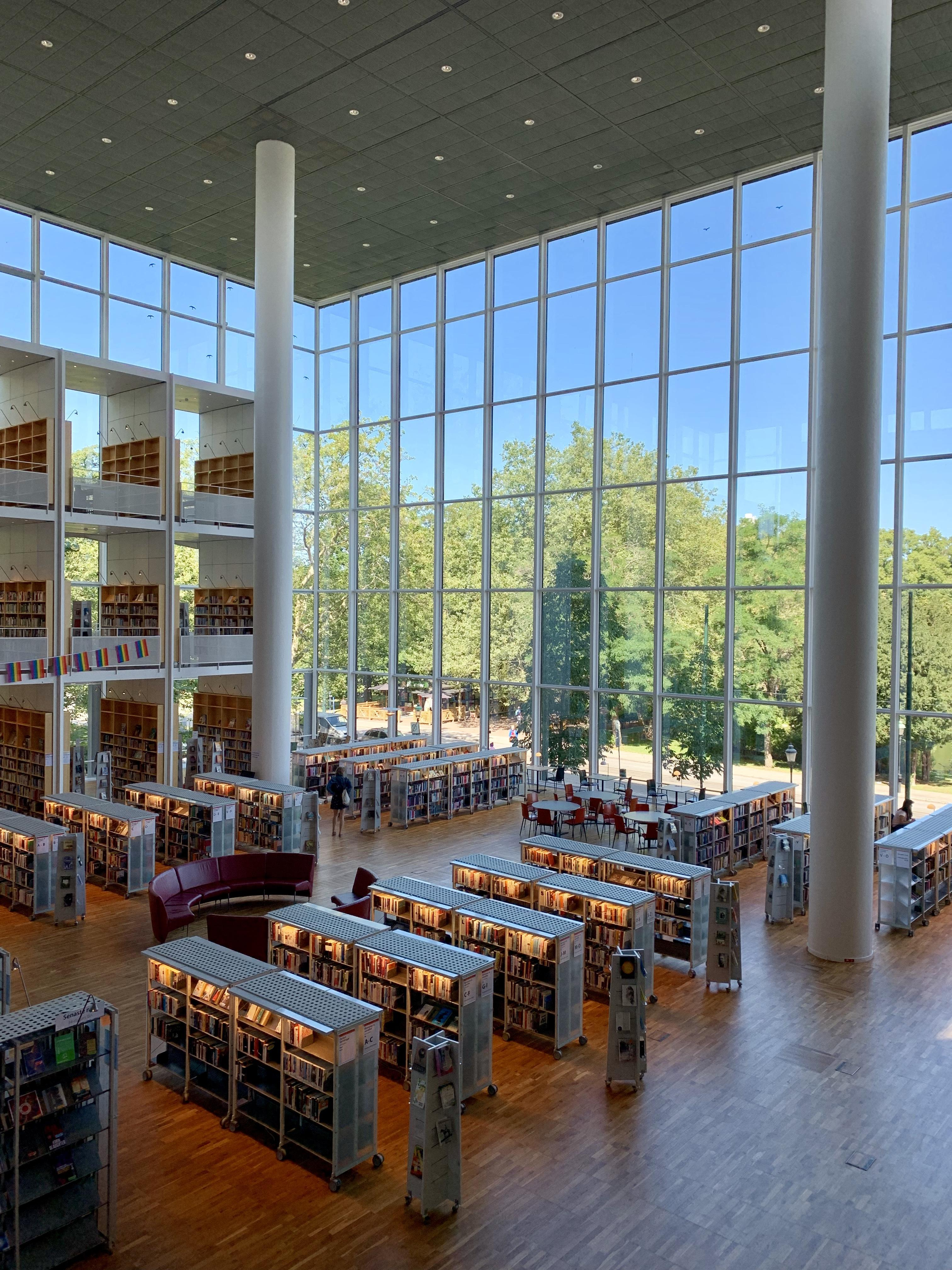}
  \end{center}
  How many rainbow flags are visible on the upper-level balcony in the image?\\
  A: 4\\
  B: 5\\
  C: 6\\
  D: 7
  
  \tcbline
  \textbf{Skill Extraction Trace} \\[5pt]
  ... The student's logic was to count the flags one by one, but they ended at 5. \texttt{...[omitted]}
  
  \tcbline
  \textbf{Extracted Skill} \\[5pt]
  \textit{\textbf{Precise enumeration of discrete objects in a sequence}}
\end{tcolorbox}

\newpage

\subsection{\dataset}
We also present qualitative examples of synthesized prompts in \dataset, focusing on skill-guided generation. Each example shows the new image passed to the data generator, the sampled skills used for conditioning, and the resulting prompt.

\vspace{10pt}
\begin{tcolorbox}[colback=white,colframe=black!75!white,colbacktitle=black!75!white,title=\dataset Example 1]
  \textbf{New Image}
  \begin{center}
  \includegraphics[width=0.80\linewidth]{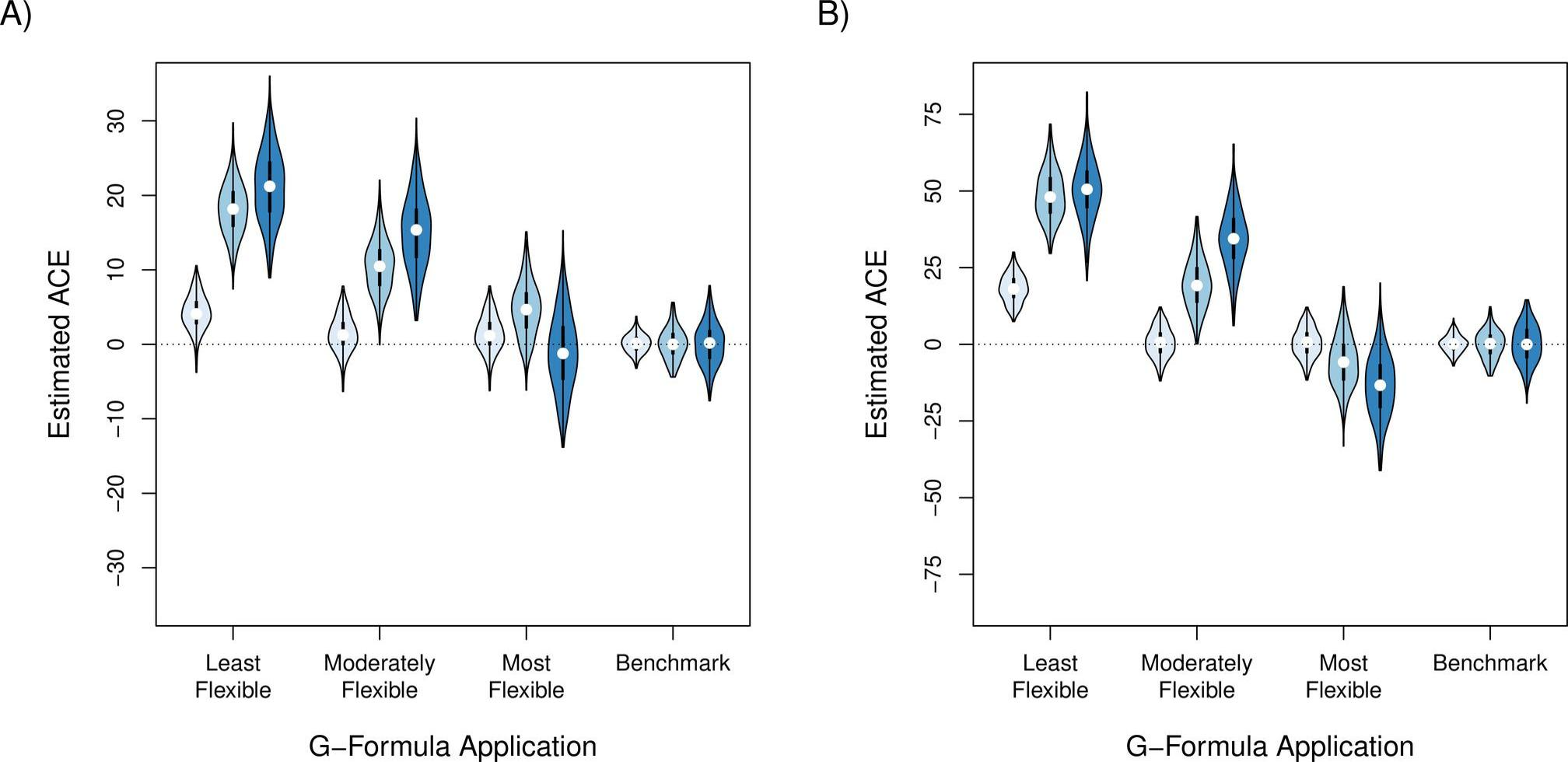}
  \end{center}

  \tcbline
  \textbf{Sampled Skills} \\[5pt]
  Correctly interpreting positions in violin plots, Multi-condition cross-validation across visual subplots, \texttt{...[omitted]}
  
  \tcbline
  \textbf{Generated Prompt} \\[5pt]
  \textit{\textbf{For Panel A, count the number of G–Formula Application categories where at least one violin plot has an Estimated ACE value exceeding 20. For Panel B, count the number of categories where at least one violin plot has an Estimated ACE value exceeding 50. What is the product of these two counts?}}
\end{tcolorbox}

\newpage

\begin{tcolorbox}[colback=white,colframe=black!75!white,colbacktitle=black!75!white,title=\dataset Example 2]
  \textbf{New Image}
  \begin{center}
  \includegraphics[width=0.54\linewidth]{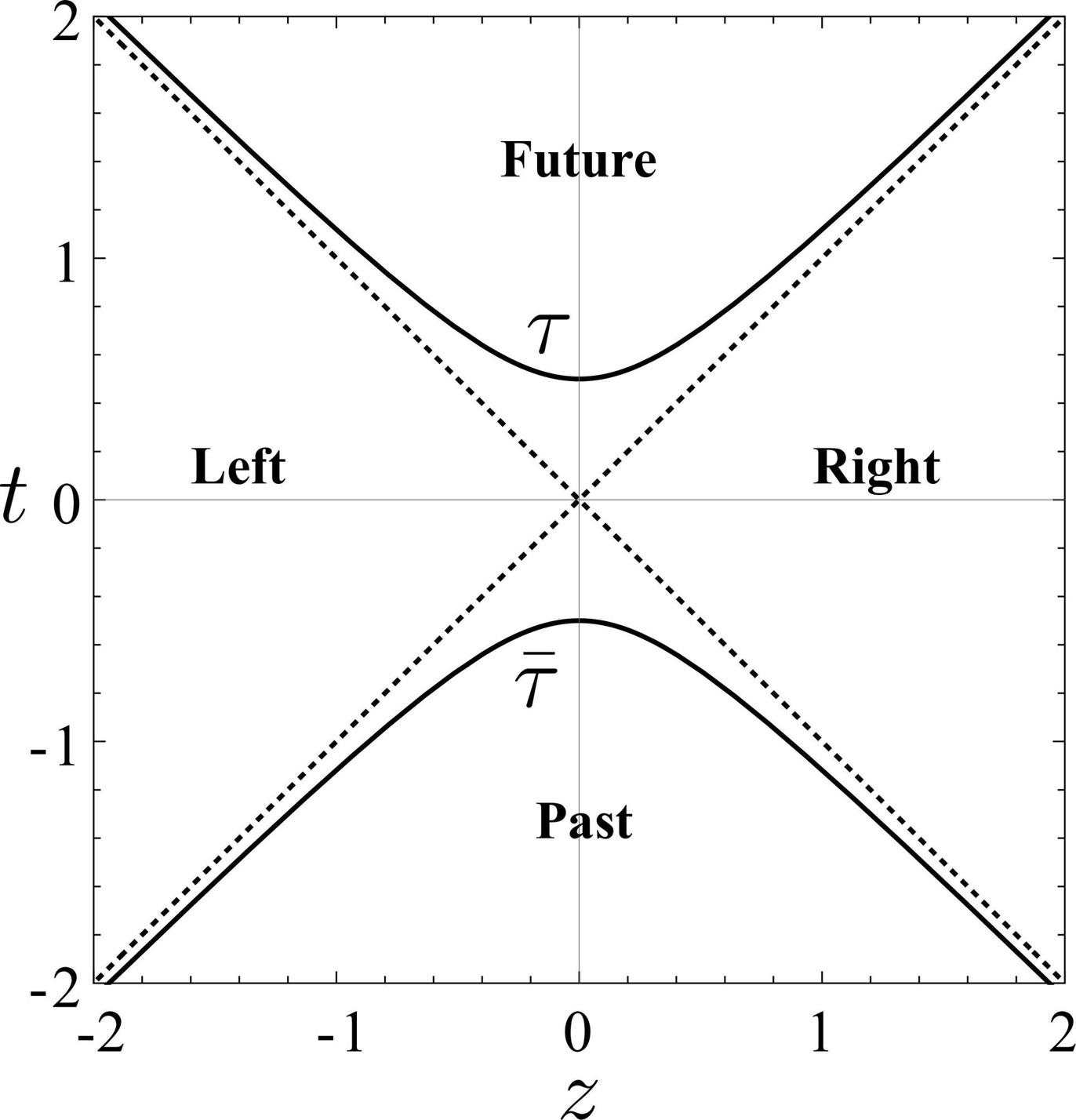}
  \end{center}

  \tcbline
  \textbf{Sampled Skills} \\[5pt]
  Visual Coordinate Extraction and Function Evaluation, Parameter Instantiation from Discrete Data Points, \texttt{...[omitted]}
  
  \tcbline
  \textbf{Generated Prompt} \\[5pt]
  \textit{\textbf{Given that the solid curve labeled $\tau$ in the Future region of the spacetime diagram represents a hyperbola defined by $t^2 - z^2 = c$ for some constant $c$, first determine $c$ using the curve's intersection with the t-axis, then calculate the slope of the curve at that intersection point.}}
\end{tcolorbox}

\begin{tcolorbox}[colback=white,colframe=black!75!white,colbacktitle=black!75!white,title=\dataset Example 3]
  \textbf{New Image}
  \begin{center}
  \includegraphics[width=0.54\linewidth]{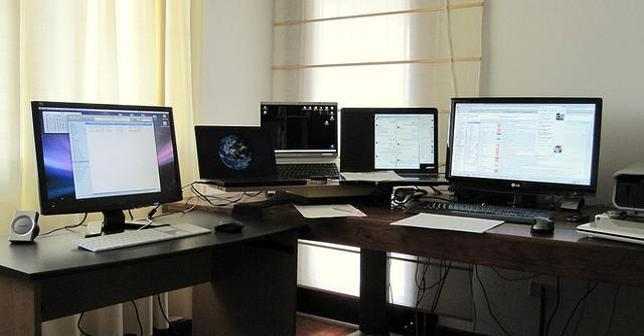}
  \end{center}

  \tcbline
  \textbf{Sampled Skills} \\[5pt]
  Coordinate system interpretation with strict inequality constraints, Accurate spatial relationship interpretation in lateral vs. depth axes, \texttt{...[omitted]}
  
  \tcbline
  \textbf{Generated Prompt} \\[5pt]
  \textit{\textbf{How many monitors are positioned such that their horizontal center lies within the rightmost 30\% of the image's width?\\
  A) 1\\
  B) 2\\
  C) 3\\
  D) 4}}
\end{tcolorbox}

\begin{tcolorbox}[colback=white,colframe=black!75!white,colbacktitle=black!75!white,title=\dataset Example 4]
  \textbf{New Image}
  \begin{center}
  \includegraphics[width=0.54\linewidth]{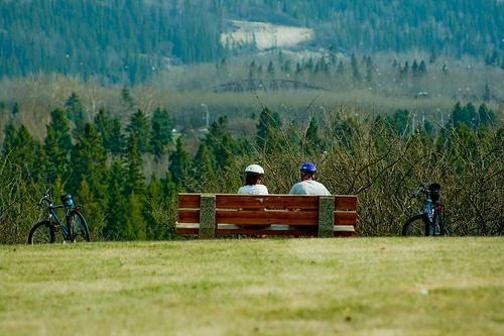}
  \end{center}

  \tcbline
  \textbf{Sampled Skills} \\[5pt]
  Color Perception Failure due to Saliency Bias, Distinguishing between object-relative and viewer-relative spatial reference frames, \texttt{...[omitted]}
  
  \tcbline
  \textbf{Generated Prompt} \\[5pt]
  \textit{\textbf{A photographer captured this scene from a position directly behind the bench. If the individual wearing the white helmet wishes to face the bicycle located on the right side of the image, in which direction must they rotate their head relative to their own body orientation?}}
\end{tcolorbox}



\end{document}